\def\year{2020}\relax
\documentclass[letterpaper]{article} % DO NOT CHANGE THIS
\usepackage{aaai20}  % DO NOT CHANGE THIS
\usepackage{times}  % DO NOT CHANGE THIS
\usepackage{helvet} % DO NOT CHANGE THIS
\usepackage{courier}  % DO NOT CHANGE THIS
\usepackage[hyphens]{url}  % DO NOT CHANGE THIS
\usepackage{graphicx} % DO NOT CHANGE THIS
\usepackage{graphicx}  % DO NOT CHANGE THIS
\title{Safe Reinforcement Learning via Probabilistic Shields}
 \author{
 Nils Jansen\textsuperscript{\rm 1}, Bettina K\"onighofer\textsuperscript{\rm 2}, Sebastian Junges\textsuperscript{\rm 3}, Alexandru C. Serban\textsuperscript{\rm 1}, Roderick Bloem\textsuperscript{\rm 2} \\
 \textsuperscript{\rm 1} Radboud University, The Netherlands\\
 \textsuperscript{\rm 2} TU Graz, Austria\\
 \textsuperscript{\rm 2} RWTH Aachen University, Germany}

\usepackage{amsmath,amssymb,amsfonts,amsthm}       % blackboard math symbols
\usepackage{booktabs}       % professional-quality tables

\usepackage{xcolor}
\usepackage{xspace}
\usepackage{todonotes}
\usepackage{paralist}
\usepackage{subfigure}
\usepackage{appendix}
\usepackage{cases}
\usepackage[pdfpagelabels=true]{hyperref}
\hypersetup{
    colorlinks=true,
    linkcolor=blue,
    filecolor=magenta,      
    urlcolor=cyan,
}
\usepackage{pgfplots}
\pgfplotsset{compat=newest}
\newtheorem{theorem}{Theorem}
\newtheorem{lemma}{Lemma}

\theoremstyle{definition}
\newtheorem{definition}{Definition}

\theoremstyle{remark}

\usetikzlibrary{fit}
\newcommand\shieldsym[1][]{%
\raisebox{-1.5pt}{\tikzset{
    shield width/.store in=\shieldwidth,
    shield width=1.5ex,
    shield height/.store in=\shieldheight,
    shield height=1.75ex
}%
\tikz [baseline,#1] \draw (0,\shieldheight) -- (0,\shieldwidth/2) arc [radius=\shieldwidth/2, start angle=-180, end angle=0] -- (\shieldwidth,\shieldheight) -- cycle;%
}}

\newcommand{\col}{C}
\newcommand{\Pos}{\ensuremath{\textsl{Pos}}}

\newcommand{\mcresmin}[2]{\ensuremath{\eta^{\min}_{#1,#2}}}
\newcommand{\mcresmax}[2]{\ensuremath{\eta^{\max}_{#1,#2}}}

\newcommand{\val}[2]{\ensuremath{\textsl{val}_{#1}^{#2}}}
\newcommand{\optval}[2]{\ensuremath{\textsl{optval}_{#1}^{#2}}}
\newcommand{\actvalues}{\ensuremath{\textsl{ActVals}}}

\newcommand{\shield}[2]{\ensuremath{\textsl{shield\,}_{#1}^{#2}}}
\newcommand{\shielded}[1]{\ensuremath{#1_{\text{\shieldsym}}}}

\newcommand{\pacman}{\textsc{PM}\xspace}

\newcommand{\Distr}{\mathit{Distr}}

\newcommand{\distDom}{X}

\newcommand{\distFunc}{\mu}
\newcommand{\distDomElem}{x}
\newcommand{\supp}{\mathit{supp}}

\newcommand{\R}{\mathbb{R}}

\newcommand{\N}{\mathbb{N}}
\newcommand{\finally}{\lozenge}
\newcommand{\mdp}{\mathcal{M}}
\newcommand{\MdpTuple}[1][]{\ensuremath{(S{#1},\Act{#1},\pmdp{#1},\rewFunction{#1})}}
\newcommand{\MdpInit}[1][]{\ensuremath{\mdp{#1}=\MdpTuple[#1]}}
\newcommand{\MdpTupleR}[1][]{\ensuremath{(S{#1},\Act{#1},\pmdp{#1})}}
\newcommand{\MdpInitR}[1][]{\ensuremath{\mdp{#1}=\MdpTupleR[#1]}}

\newcommand{\act}[1][a]{\alpha} % single action of MDP
\newcommand{\Act}{\mathit{Act}}        % action set of MDP

\newcommand{\sched}{\ensuremath{\sigma}}

\newcommand{\rewFunction}{\ensuremath{{r}}}

\newcommand{\pmdp}{\mathcal{P}}

\newcommand{\ltp}[1][L]{\mathcal{L}}   % linear-time property
\newcommand{\eg}{e.\,g.\xspace}

\newcommand{\bddstate}[1]{s_{\mathbb{B}}{}}

\newcommand{\tool}[1]{\texttt{#1}\xspace}

\newcommand{\prism}{\tool{PRISM}}
\newcommand{\storm}{\tool{Storm}}

\DeclareRobustCommand{\cpp}
{\valign{\vfil\hbox{##}\vfil\cr
   \textsf{C\kern-.1em}\cr
   $\hbox{\fontsize{\sf@size}{0}\textbf{+\kern-0.05em+}}$\cr}%
}

\selectcolormodel{cmy}

\definecolor{lgrey}{rgb}{0.8,0.8,0.8}
\definecolor{grey}{rgb}{0.5,0.5,0.5}

\definecolor{lightblue}{rgb}{0.8,0.8,1.0}
\definecolor{lightred}{rgb}{1.0,0.8,0.8}
\definecolor{lightgreen}{rgb}{0.8,1.0,0.8}
\definecolor{angrygreen}{cmyk}{0.279,0,0.91,0.08}
\definecolor{lightred}{rgb}{1.0,0.8,0.8}
\definecolor{pink}{rgb}{1.0,0.1,1.0}

\definecolor{prismgreen}{HTML}{009900}
\definecolor{prismred}{HTML}{cc0000}
\definecolor{prismblue}{HTML}{0000cc}

\begin{document}

\maketitle

%\urlstyle{tt}

\begin{abstract}
	This paper targets the efficient construction of a safety shield for decision making in scenarios that incorporate uncertainty.
Markov decision processes (MDPs) are prominent models to capture such planning problems.
Reinforcement learning (RL) is a machine learning technique to determine near-optimal policies in MDPs that may be unknown prior to exploring the model.
However, during exploration, RL is prone to induce behavior that is undesirable or not allowed in safety- or mission-critical contexts.
We introduce the concept of a probabilistic shield that enables decision-making to adhere to safety constraints with high probability. 
In a separation of concerns, we employ formal verification to efficiently compute the probabilities of critical decisions within a safety-relevant fragment of the MDP.
We use these results to realize a shield that is applied to an RL algorithm which then optimizes the actual performance objective.
We discuss tradeoffs between sufficient progress in exploration of the environment and ensuring safety.
In our experiments, we demonstrate on the arcade game PAC-MAN and on a case study involving service robots that the learning efficiency increases as the learning needs orders of magnitude fewer episodes.

%We demonstrate tradeoffs between sufficient progress in exploration of the environment and ensuring strict safety.

%The construction of the shield exploits the inherent uncertainties in scenarios given by Markov decision processes.

%A prominent problem in artificial intelligence and machine learning is the safe exploration of an environment.
%In particular, reinforcement learning is a well-known technique to determine optimal policies for complicated dynamic systems, but suffers from the fact that such policies may induce harmful behavior.
%We present the concept of a shield that forces decision-making to provably adhere to safety requirements with high probability.
%Our method exploits the inherent uncertainties in scenarios given by Markov decision processes.
%We present a method to efficiently compute the probabilities of critical decisions regarding temporal logic constraints.
%We use that information to realize a shield that---when applied to a reinforcement learning algorithm---ensures (near-)optimal behavior both for the safety constraints and for the actual learning objective.
%In our experiments, we show on the arcade game PAC-MAN that the learning efficiency increases as the learning needs orders of magnitude fewer episodes.
%We demonstrate tradeoffs between sufficient progress in exploration of the environment and ensuring strict safety.
%

\end{abstract}

\section{Introduction}
%\paragraph{General Motivation.}

%In recent years, artificial intelligence (AI), and in particular reinforcement learning (RL)~\cite{sutton1998reinforcement}, evolved from areas like game-playing or language-translation to critical domains such as health, energy, defense, or transportation.
%A major open challenge is the \emph{safety} of decision-making for systems employing RL~\cite{stoica2017berkeley,freedman2016safety,future_AI,russel_priorities}, in particular during the exploitation phase.
%%\nj{removed: One reason is that restricting decisions to those that are safe is not always assessable by defining simple rules, especially during runtime.}
%The area of \emph{safe exploration} aims to guide RL agents to adhere to safety requirements during this phase~\cite{pecka2014safe,amodei2016concrete}.

Recent years showed increased use of reinforcement learning (RL) in solving tasks such as complex games~\cite{silver2016mastering} or robotic manipulation~\cite{wang2019learning}.
% or safety critical applications~\cite{shalev2016safe}.
In RL, an agent perceives the surrounding environment and acts towards maximizing a long-term reward signal.
A major open challenge is the \emph{safety} of decision-making for systems employing RL~\cite{stoica2017berkeley,freedman2016safety}.
% in particular during the exploration phase.
%Moreover, in safety-critical tasks the agent is also constrained to avoid unintended or harmful behavior.
Particularly during the exploration phase, when an agent chooses random actions in order to examine its surroundings, it is important to avoid actions that may cause unsafe outcomes.
The area of \emph{safe exploration} investigates how RL agents can adhere to safety requirements during this phase~\cite{pecka2014safe,amodei2016concrete}.

%One suitable technique that delivers theoretical guarantees are \emph{safety-shields}~\cite{DBLP:conf/tacas/BloemKKW15,shield_rl}, which can be interpreted as teachers that prevent the learner from unsafe decisions at runtime.
%To that end, the optimality criterion (the performance objective) is extended with a strict constraint that unsafe system states are \emph{never} visited, i.e., there are no \emph{safety violations} (the safety objective).
%However, so far, shields do not take randomness into account and are amenable only to systems with policies that avoid safety violations all together.
%Policies satisfying such strict constraints may not exist.

One suitable technique that delivers theoretical guarantees are so-called \emph{safety-shields}~\cite{DBLP:conf/tacas/BloemKKW15,shield_rl}.
Shields prevent an agent from taking unsafe actions at runtime.
To this end, the performance objective is extended with a constraint specifying that unsafe states should \emph{never} be visited.
This new safety objective ensures there are no violations during the exploration phase.
So far, shields have showed success in deterministic settings, where an agent avoids safety violations altogether.
However, in many cases this tight restriction limits the agent's exploration and understanding of the environment, and policies satisfying the restrictions may not even exist.

We propose to incorporate more liberal constraints that enforce safety violations to occur \emph{only with small probability}.
%\textcolor{blue}{Each time the agent is about to make a
%decision, the shield computes for each action available the minimal probability for violating the safety specification after taking this action.
%We base our notion of a shield on Markov decision processes (MDPs) which constitute a popular modeling formalism for decision-making under uncertainty~\cite{thrun2005probabilistic}.
%%, and underlie the standard formulations of RL.
%We assess safety by means of probabilistic \emph{temporal logic constraints}~\cite{BK08} that limit, for example, the probability to reach a set of critical states in the MDP.
If an action increases the probability of a safety violation by more than a threshold $\delta$ with respect to the optimal safety probability, the shield blocks the action from the agent.

Consequently, an agent augmented with a shield is \emph{guided} to satisfy the safety objective during exploration (or as long as the shield is used).
The shield is \emph{adaptive} with respect to $\delta$,
as a high value for $\delta$ yields a stricter shield, a smaller value a more permissive shield.
The value for $\delta$ can be changed on-the-fly, and may depend on the individual minimal safety probabilities at each state.
Moreover, in case there is not suitable safe action with respect to $\delta$, the shield can always pick the optimal action as a fallback.

We base our formal notion of a probabilistic shield on MDPs, which constitute a popular modeling formalism for decision-making under uncertainty~\cite{white1985real} and is widely used in model-based RL.
We assess safety by means of probabilistic \emph{temporal logic constraints}~\cite{BK08} that limit, for example, the probability to reach a set of critical states in the MDP.
%That is, the learner is guaranteed to take actions that are close to the optimal action regarding the safety objective.
%Such safety constraints are typical in technical standards and captured by notions of safety integrity levels---e.g.~\cite{iso26262,ISO13849,IEC:61508}---and natural in presence of lossy communication channels~\cite{Suriyachai:2012,Martelli:2012}, transient faults~\cite{vN56,DBLP:journals/pieee/HaselmanH10}, or mechanical wear~\cite{vesely1981fault}.
%Our weaker constraints are realizable in many more settings.
%
%The shield restricts the actions that an agent may take to those that satisfy the safety constraints.
%Consequently, an agent augmented with a shield is \emph{guided} to satisfy the safety objective during exploration (or as long as the shield is used).
%This induces a lower probability of safety violations by the augmented learner agent to adhere to safety specifications, and may additionally improve performance.
%

In order to assess the risk of one action, we (1) construct a behavior model for the environment using model-based RL~\cite{dayan2008reinforcement}. %, in a training environment together with suitable data augmentation techniques.
We can plug this model into any concrete scenario to obtain an MDP.
%for the shield computation.
To construct the shield, we (2) use a model-based verification technique known as \emph{model checking}~\cite{DBLP:books/daglib/0007403,BK08} that assesses whether a system model satisfies a specification.
Due to its rigor, the validity of results \emph{only} depends on the quality of the model, and
%We use (2) probabilistic model checking~\cite{Kat16,DBLP:conf/lics/Kwiatkowska03},
%featuring off-the-shelf tool support~\cite{DBLP:conf/cav/KwiatkowskaNP11,DJKV17} and yielding
we obtain precise \emph{safety probabilities of any possible decision} within the MDP.
These probabilities can be looked up efficiently and compared to the %safety 
threshold $\delta$.
The shield then readily (3) augments either model-free or model-based RL.
%, assuming the model is adequate.
%

We identify three key challenges:

Firstly, model checking~--~as any model-based technique~--~is susceptible to scalability issues.
A key advantage of using a separate safety objective is that we may analyze safety on just a fraction of the system, the \emph{safety-critical MDP}.
In our experiments, these MDP fragments are at least ten orders of magnitude smaller than a full model of the system, rendering model checking applicable to realistic scenarios.
%We stress that the learner respecting the performance has to consider the full model, but may do so using either model-based or model-free approaches.
We introduce further optimizations based on problem-specific abstraction techniques.

Secondly, without randomness, all states are either absolutely safe or unsafe.
However, in the presence of randomness, safety may be seen as a quantitative measure: in some states all actions may induce a large risk, while one action may be considered \emph{relatively} safe.
Therefore, it is essential to have an \emph{adaptive} notion of shielding, in which the pre-selection of actions is not based on absolute thresholds.

Lastly, shielding may \emph{restrict} exploration and lead to suboptimal policies. Therefore, it should not be considered in isolation.
The trade-off between optimizing the performance objective and the achieved safety is intricate.
Intuitively, accepting small short term risks may allow for efficient exploration and limit the risk long-term.
To this end, we provide and discuss mechanisms that allow to adjust the shield based on such observations.

%of the approach are introduced.
We apply shielding to two distinct use cases: the arcade game PACM-MAN and a new case study involving service robots in a warehouse.
Shielded RL leads to improved policies for both case studies with fewer safety violations and performance superior to unshielded RL.
\smallskip

Supplementary materials are available at \url{http://shieldrl.nilsjansen.org}.

\paragraph{Related Work.}
%Alongside related work mentioned above...
%During the exploration of the MDP, the current policy may be unsafe in the sense that it harms the agent or the environment.
%This shortcoming restricts the application of RL mainly to
%academic or uncritical
%application areas where safety is not a concern and has triggered the direction of safe exploration for RL, in short \emph{safe RL}~\cite{garcia2015comprehensive,pecka2014safe}.
Most approaches to safe RL~\cite{garcia2015comprehensive,pecka2014safe} rely on reward engineering and effectively changing the learning objective.
%\nj{more references, focus on NIPS, ICML, AAAI, IJCAI. consider related work and paragraphs from~\cite{shield_rl}}
%, while our goal is to incorporate formal verification to ensure safety specifications.
In contrast to ensuring temporal logic constraints, reward engineering designs or ``tweaks'' the reward functions such that a learning agent behaves in a desired, potentially safe, manner.
As rewards are specialized for particular environments, reward engineering runs the risk of triggering negative side effects or hiding potential bugs~\cite{sculley2014machine}.
Recently, it was shown that reward engineering is not sufficient to capture temporal logic constraints in general~\cite{DBLP:conf/tacas/HahnPSSTW19}.
Additionally, in~\cite{cheng2019end} the exploration of model-free RL algorithms is limited using control barrier functions and in~\cite{garcia2019probabilistic} exploration is restricted to a space close to an optimal, precomputed policy.

First approaches directly incorporating formal specifications tackle this problem with pre-computations; making assumptions on the available information about the environment~\cite{DBLP:conf/iros/WenET15,junges-et-al-tacas-2016,fulton2018safe,hasanbeig2018logically,DBLP:conf/icaart/MasonCKB17,moldovan2012safe}, by employing PAC guarantees~\cite{Fu-RSS-14}, or by an intermediate ``correction'' of policies~\cite{DBLP:conf/nfm/PathakAJTK15}.
Most related is~\cite{shield_rl}, which introduces the concept of a shield for RL.
%In contrast to our work, however, the stochastic behavior of the MDP is ignored. Consequently, the learning agent does not take any risks, which is unrealistic in most scenarios.
The difference and novel contribution is rooted in the consideration of stochastic behavior, which is natural to RL.
Intuitively, without stochasticities, a learning agent does not take any risk, which is unrealistic in most scenarios.
Moreover, often one cannot assume that a $100\%$ (or almost-sure) safety is realizable.
A similar approach to ours was developed independently in~\cite{DBLP:journals/corr/abs-1904-07189}, but targets a different case study and does not consider scalability issues of formal verification.
In a related direction, methods from reinforcement learning have been successfully employed to improve the scalability of verification methods for MDPs.
Such approaches often use rich specifications like $\omega$-regular languages as a control to guide the exploration of MDP during learning~\cite{sadigh2014learning,DBLP:conf/atva/BrazdilCCFKKPU14,hasanbeig2018logically,DBLP:conf/concur/KretinskyPR18,DBLP:journals/corr/abs-1810-00950}.
%For instance~\cite{DBLP:conf/atva/BrazdilCCFKKPU14} provides upper and lower bounds on safety at all times.

Safe model-based RL for continuous state spaces employing Lyapunov functions is considered in~\cite{berkenkamp2017safe,chow2018lyapunov}.
 \tool{UPPAAL STRATEGO} provides a number of algorithms combining safety synthesis with optimizing RL for continuous space MDPs~\cite{DBLP:conf/tacas/DavidJLMT15}.
 Finally,~\cite{ohnishi2018safety} uses control barrier functions (CBFs) for safe RL.

Probabilistic planning considers similar problems as probabilistic model checking~\cite{DBLP:journals/jair/SteinmetzHB16,kolobov2012planning}.
A recent comparison between tools from both areas can be found in \cite{DBLP:conf/tacas/HahnHHKKKPQRS19}.

%
%\section{Foundations}
\section{Problem Statement}\label{sec:problem}

%\subsection*{Background}

\paragraph{Foundations.}
A \emph{probability distribution} over a countable set $\distDom$ is a function $\distFunc\colon\distDom\rightarrow[0,1]$ with $\sum_{\distDomElem\in\distDom}\distFunc(\distDomElem)=1$.
$\Distr(\distDom)$ denotes all distributions on $\distDom$. The support of $\distFunc\in\Distr(\distDom)$ is $\supp(\distFunc)=\{x\in\distDom \mid \distFunc(x){>}0\}$.
%
%%\paragraph{MDPs.}
%\begin{definition}[MDP]
A \emph{Markov decision process} (MDP) $\MdpInit$ has a set $S$ of \emph{states}, a finite set $\Act$ of \emph{actions}, a (partial) \emph{probabilistic transition function} $\pmdp\colon S\times\Act\rightarrow\Distr(S)$, and an \emph{immediate reward function} $\rewFunction \colon S \times \Act \rightarrow \R_{\geq 0}$.
For all $s\in S$ the available actions are $\Act(s)=\{\act\in\Act\mid \pmdp(s,\act) \neq \bot \}$ and we assume $|\Act(s)|\geq 1$.
%\end{definition}
%
%MDPs operate by means of nondeterministic \emph{choices} of actions at each state.
%The probability distribution for a choice determines probabilistically the successor state.
%The reward function may be omitted.
%We disallow deadlock states: $|\Act(s)|\geq 1$ for all $s\in S$.
%If $|\Act(s)|=1$ for all $s\in S$, actions are superfluous and the MDP $\mdp$ reduces to a \emph{discrete-time Markov chain (MC)}
%, sometimes denoted by $\dtmc$, with a  transition probability function $\pmdp\colon S\rightarrow\Distr(S)$.
%
%\begin{definition}[Policy]
A \emph{policy} is a function $\sched\colon S^*\rightarrow\Distr(\Act)$ with $\supp(\sigma(s_1\hdots s_n)) \subseteq \Act(s_n)$ and $S^*$ a finite sequence of states.	
%\end{definition}
%For many specifications, it suffices to consider \emph{stationary}, \emph{deterministic} policies $\sched\colon S \rightarrow \Act$~\cite{Put94}.
%For multiple---possibly conflicting---specifications, more general policies (with randomization and finite memory) are necessary~\cite{DBLP:conf/stacs/ChatterjeeMH06}.
%The application of a policy $\sched$ to an MDP $\mdp$ yields an \emph{induced MC}.

%\paragraph{Temporal logic and Probabilistic Model Checking}

In formal methods, safety properties are often specified as \emph{linear temporal logic} (LTL) properties~\cite{pnueli1977temporal}.
% or computation-tree logic (CTL)~\cite{clarke1981design}.
%
%\emph{Model checking} is a fully automatic verification technique~\cite{DBLP:books/daglib/0007403,BK08} that assesses for a system model wether it satisfies a specification.
%Due to its rigor, the reliability of the results \emph{only} depends on the quality of the model.
%Based on a system model and a specification, model checkers automatically assess whether the model satisfies the properties.
%In contrast to testing and simulation, model checking is not biased to certain scenarios, but instead relies on an exhaustive state-space exploration.
%As for any model-based technique, the reliability of verification results relies on the correctness of the system model.
%
%
%We consider a variant of temporal logic called \emph{probabilistic computation tree logic} (PCTL)~\cite{hansson1994logic}.
For an MDP $\mdp$, probabilistic model checking~\cite{Kat16,DBLP:conf/lics/Kwiatkowska03}
 employs value iteration or linear programming to compute the probabilities of \emph{all states and actions of the MDP}
% $\pr(\varphi)$
 to satisfy an LTL property $\varphi$.
% As an example, we may be interested in a policy for an POMDP such that $\pr(\finally T)\geq \lambda\in[0,1]$, that is, if the probability to ``eventually'' reach a set of target states $T\subseteq S$ is above a threshold.
%
%such a specification is for instance of the form $\reachPropuT$, which is satisfied for $\mdp$, if the probability to ``eventually'' reach a set of target states $T\subseteq S$ is at least $\lambda\in[0,1]$, when considering all possible policies.
%The formula $\finally T$ is called a path formula, $\reachPropuT$ is called a state formula.
%%In our setting, we require satisfaction of the state formula at every MDP state.
%%
%Probabilistic model checking~\cite{Kat16,DBLP:conf/lics/Kwiatkowska03} %,DBLP:series/natosec/Baier16
% employs methods based on, e.g., value iteration or linear programming.
%  to verify such specifications for MDPs.
%,
% or \tool{Modest}~\cite{DBLP:conf/tacas/HartmannsH14}.
Specifically,
%for MDP $\mdp$ and specification $\p_{\geq \lambda}(\varphi)$,
 we compute $\mcresmax{\varphi}{\mdp} \colon S \rightarrow [0,1]$ or $\mcresmin{\varphi}{\mdp} \colon S \rightarrow [0,1]$,
%where $\mcresmin{\varphi}{\mdp}(s)$ (or $\mcresmax{\varphi}{\mdp}(s)$)
which give for all states the minimal (or maximal) probability over all possible policies to satisfy $\varphi$.
For instance,  for $\varphi$ encoding to reach a set of states $T$,  $\mcresmax{\varphi}{\mdp}(s)$ describes the maximal probability to ``eventually'' reach a state in $T$.
% set of target states $T\subseteq S$ from state $s\in S$.

\paragraph{Setting.}

We define a setting
%partially-controlled multi-agent system~\cite{brafman1996partially,van2008multi},
where one controllable agent (the \emph{avatar}) and a number of uncontrollable agents (the \emph{adversaries}) operate within an \emph{arena}.
The arena is a compact, high-level description of the underlying model.
From this arena, the potential states and actions of all agents may be inferred.
For safety considerations, the %token-based %BK-NOTE: removed token, because we mention it here for the first time and I think it is better to explain it later properly
reward structure can be neglected, effectively reducing the state space for our model-based safety computations.
% and discuss several potential applications.
%In particular, we define a finite graph representation of an arena in which agents operate.
%where \textbf{agents} move within a discrete \textbf{arena}.
%\begin{definition}
Formally, an \emph{arena} is a directed graph $G = (V,E)$ with a finite sets $V$ of nodes and $E\subseteq V\times V$ of edges.
The agent's \emph{position} is defined via the current node $v\in V$. The agent \emph{decides} on a new edge $(v, v') \in E$ and determines its next position $v'$.

Some (combinations of) agent positions are safety-critical, as they e.g.\ correspond to collisions or falling off a cliff. %\nj{trap states, cliff}
A safety property may describe reaching such positions, or use any other property expressible in (the safety fragment of) temporal logic.

While the underlying model for the arena suffices to specify the safe behavior, it is not sufficiently succinct to model the performance via rewards. 
Consider an edge that is safety-relevant, but the agent is only rewarded the first time taking this edge.
Thus, in a flat model with rewards, two different edges are necessary to model this behavior. 
However, the reward (and thus the difference between these edges) is not needed to assess the safety, and the safety-relevant model may be pruned to an exponentially smaller model.
We use a \emph{token} function that implicitly extends the underlying model by a reward structure, enabling a separation of concerns between safety and performance.
%
%with an implicit exponential blow-up, such that the states are adequately annotated with rewards.

Technically, we associate edges with a  token function $\circ \colon E\rightarrow\{0,1\}$, indicating the status of an edge.
%Naturally, tokens can be extended to describe an $n$-ary status.
Tokens can be (de-)~activated and have an associated \emph{reward} earned upon taking edges with an active token.

\emph{Example 1: Autonomous driving.}
An autonomous taxi (the avatar) operates within a road network encoded by an arena.
	The taxi has to visit several points to pick up or drop off passengers~\cite{dietterich2000hierarchical,taxi}.
 	Upon visiting such a point, a corresponding token activates and a reward is earned, afterwards the token is deactivated permanently.
 	Meanwhile, the taxi has to account for other traffic participants or further environmental factors (the adversaries).
 	A sensible safety specification may restrict the probability for collision with other cars to $0.5\%$.
 	Note that the token structure is not relevant for such a specification.
 	% 	, for which one may have learned behavior models over time~\cite{sadigh2016planning,sadigh2018planning}.
%	By means of our shielding technique,	 we can achieve \emph{safe} learning to obtain an optimal route for the car.
%\end{example}

\emph{Example 2: Robot logistics in a smart factory.}
Take a factory floor plan with several corridors with machines.
The adversaries are (possibly autonomous) transporters moving parts within the factory.
The avatar models a specific service unit moving around and inspecting machines where an issue has been raised (as indicated by a token), while accounting for the behavior of the adversaries.
 Corridors might be to narrow for multiple (facing) robots, which poses a safety critical situation.
 The tokens allow to have a \emph{state-dependent} cost, either as long as they are present (indicating the costs of a broken machine) or for removing the tokens (indicating costs for inspecting the machine).
 A similar scenario has been investigated in~\cite{DBLP:journals/corr/abs-1806-07135}.

\paragraph{Problem.}%
Consider an environment described by an arena as above and a safety specification.
We assume stochastic behaviors for the adversaries, e.g, obtained using RL~\cite{sadigh2018planning,sadigh2016planning} in a training environment.
In fact, this stochastic behavior determines all actions of the adversaries via probabilities.
The underlying model is then a Markov decision process: the avatar executes an action, and upon this execution the next exact positions (the state of the system) are determined stochastically.

We compute a $\delta$-shield that prevents avatar decisions that violate this specification by more than a threshold $\delta$ with respect to the optimal safety probability.
We evaluate the shield using a model-based or model-free RL avatar that aims to optimize the performance.
The shield therefore has to handle an intricate tradeoff between strictly focussing on (short and midterm) safety and performance.

\section{Constructing Shields for MDPs}
\label{sec:shields}
\begin{figure*}[t]
\centering
\scalebox{0.75}{
\input{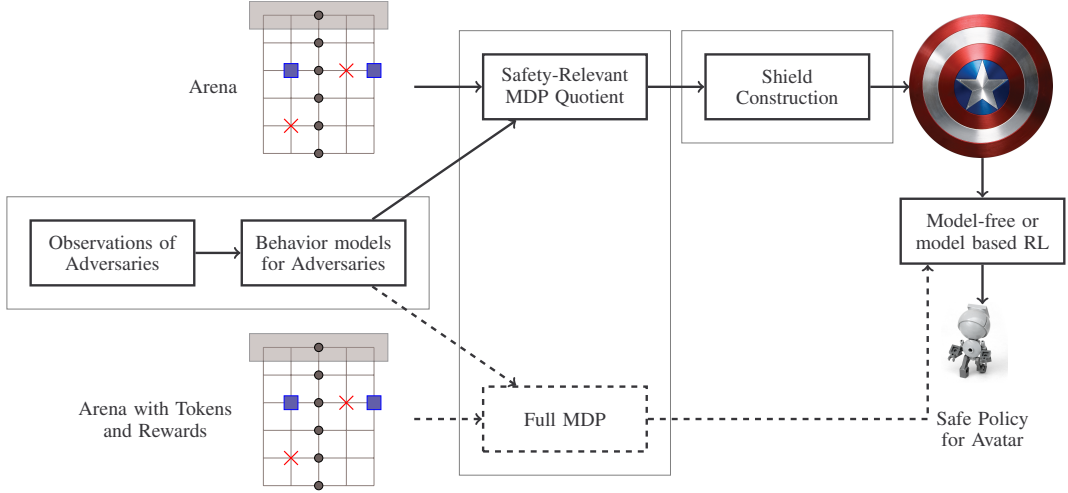}
}
\caption{Workflow of the Shield Construction}
\label{fig:flowchart}
\end{figure*}

%We consider a scenario where model-based or model-free RL aims to maximize the expected reward within an MDP, while unsafe actions are blocked by a probabilistic shield.
We outline the workflow of our approach in Fig.~\ref{fig:flowchart} and below.
%, starting with the setting from the previous section.
%First, based on observations in multiple arenas, we construct a general (stochastic) \textbf{behavior model} for each adversary.
%Combining these models with a concrete arena yields an MDP.
%At this point, we ignore the token function (and necessarily the potentially unknown reward function), so the MDP is a \textbf{safety-relevant quotient} of the full model.
%% that models the real system within which we only assess safe behavior.
%%We therefore call the MDP the \textbf{safety-relevant quotient}.
%The underlying \textbf{full MDP} incorporates the \textbf{token function}, where associated rewards may be known or observed during learning.
%Including tokens constitutes an \textbf{exponential blowup of the safety-relevant quotient}.
%% rendering probabilistic model checking or planning practically infeasible.
%Using the safety-relevant MDP, we construct a \textbf{shield} using probabilistic model checking.
%
%As we augment the RL with the pre-computed shield, all decisions are \textbf{guaranteed to form an overall policy that adheres to safety requirements}.\nj{weaken here}
%\paragraph{Separation of Concerns: Probabilistic Model Checking and Reinforcement Learning.}
 We employ a separation of concerns between the model-based shield construction and potentially model-free reinforcement learning (RL).
First, we construct a \emph{behavior model} for each adversary.
Based on this model and a concrete arena, we construct a compact MDP model: the \emph{safety-relevant MDP quotient}.
% of the full MDP that incorporates the token function and the potentially unknown reward function.
% that models the real system within which we only assess safe behavior.
%We therefore call the MDP the \textbf{safety-relevant quotient}.
%The underlying \textbf{full MDP} incorporates the \textbf{token function}, where associated rewards may be known or observed during learning.
%Including tokens constitutes an \textbf{exponential blowup of the safety-relevant quotient}.
% rendering probabilistic model checking or planning practically infeasible.
In this MDP, we compute the \emph{shield} which enables safe RL for the full MDP.
%Then, we determine optimal behavior using reinforcement learning for a large scenario with the a-priori unknown performance specification.
% have to separate the problem to make use of the individual strengths of the two disciplines.
%By learning adversary behavior for general scenarios, we enable to
%We first construct a compact MDP model in which we can assess safety.
%Only by neglecting the token and the reward structure, we are able to employ model checking (or probabilistic planning) in a feasible way.
%Secondly, we do not construct the full MDP explicitly, but rather utilize RL's power of dealing with large-scale scenarios.
We now detail the individual technical steps to realize our proposed method.

\paragraph{Behavior Models for Adversaries.}\label{sec:learning_adv}
We learn an adversary model by observing behavior in a set of similar (small) arenas, until we gain sufficient confidence that more training data would not change the behavior significantly~\cite{sadigh2018planning}.
An upper bound on the necessary data may be obtained using Hoeffding's inequality~\cite{ziebart2008maximum}.
To reduce the size of the training set, we devise a data augmentation technique using domain knowledge of the arenas~\cite{krizhevsky2012imagenet,witten2016data}.
In particular, we abstract away from the precise configuration of the arena by partitioning the graph into zones that are relative to the view-point of the adversary (\eg, near or far, north or south, east or west).
The intuitive assumption is that the specific position of an adversary is not important, but some key information is (e.g., the relation to the position of the avatar).
This approach (1)~speeds up the learning process and (2)~renders the resulting behavior model applicable for varying the concrete instance of the same setting.

Zones are uniquely identified by a coloring with a finite set $\col$ of colors.
Formally, for an arena $G = (V,E)$, \emph{zones relative to a node $v\in V$} are given by a function $z_v\colon V \rightarrow \col$.
%\end{definition}
%
For nodes $x,y \in V$, with $z_v(x) = z_v(y)$, the assumption is that the adversary in $v$ behaves similarly regardless whether the avatar is in $x$ or $y$.
From our observations, we extract a \emph{histogram} $h\colon E \times \col \rightarrow \N$, where $h(e,c)$ describes how often the adversary takes an edge $e = (v,v')\in E$ while the  avatar is in a node $u$ with $z_v(u) = c$.
%For $h(e,c)=n$, the likelihood $n\in\N$ depends on the color $z(v)=c$ for the source node of the edge $e=(v,v')$.\nj{use another symbol for likelihood}
%In particular, for all edges $e=(v_1,v_1')$ and $e'=(v_2,v_2')$ with $z(v_1)=z(v_2)=c$, it holds that $h(e,c)=h(e',c)$.\nj{is that true?}
We translate these likelihoods into distributions over possible edges in the arena.
% to obtain the behavior model for adversaries.
\begin{definition}[Adversary Behavior]
For an arena $G = (V,E)$, zones $z_u\colon V \rightarrow \col$ for every $u\in V$, and a histogram $h\colon E \times \col \rightarrow \N$, the \emph{adversary behavior} is a function $B\colon V \times \col \rightarrow \Distr(E)$ with
\[
	B(v,c)=\frac{h\big((v,v'),c\big)}{\sum_{(v,v')\in E} h\big((v,v'),c\big)}\ .
\]	
\end{definition}
While we employ a simple normalization of likelihoods, alternatively one may also utilize, \eg, a softmax function which is adjustable to favor more or less likely decisions~\cite{sutton1998reinforcement}.

%\subsection{Constructing the Shield}\label{sec:shield}

\paragraph{Safety-Relevant Quotient MDP.}
%We describe how to construct the safety-relevant (quotient) MDP we employ for safety analysis.
The construction of the MDP $\MdpInitR$ augments an arena by behavior models $B_i$.
%  for adversaries.
First, the \emph{states} $S=V^{m+1} \times \{0,\hdots, m\}$ encode the positions for all agents and whose turn it is.
The \emph{decision states} of the safety-relevant MDP $\mdp$ are $S_d=\{s_d\in S \mid s_d=(\ldots,0)\}$,
i.e., it's the turn of the avatar.
The \emph{actions} $\Act = \{ \act_0 \} \cup \Act_E$ with $\Act_E = \{ \act_e \mid e \in E \}$ determine the movements of the avatar and the adversaries.
% in the following way.
For $(v, \hdots, 0)=s_d \in S_d$ (the avatar moves next), the available actions are $\act_e\in\Act(s_d)\subseteq\Act_e$, where $\act_e$ corresponds to an outgoing edge of $v$. For $(v, \hdots, 0)=s_d\in S_d$, $\act_e$ with $e=(v,v')$ leads with probability one to a state $s_e =(v', \hdots, 1)$.
For $(v, \hdots, v_i, \hdots, i >0)$ (an adversary moves next), there is a unique action $\act_0$ where $v_i$ is changed to $v'_i$, randomly determined according to the behavior $B_i$, which also updates $i$ to $i+1$ modulo $m$.
These transitions induce the only probabilistic choices in the MDP.

A policy only has to choose an action at decision states.
At all other states, only the unique action $\act_0$ emanates.
Consequently, a policy for $\mdp$ is a policy for the avatar.
%}

%\paragraph{Constructing The Full MDP.}
 In theory, one can build the full MDP for the arena $(V,E)$ and the token function $\circ \colon E\rightarrow\{0,1\}$ under the assumption that the reward function is known.
Then, one can compute the reward-optimal and safe policy without need for further learning techniques.
As there are $2^E$ token configurations, the state space blows up exponentially, which prevents the successful application of model checking or planning techniques for anything but very small applications.

%The \textbf{states} $S=\Pos^{m+1} \times \{0,\hdots, m\}$ encode the positions for all agents.
%The \textbf{actions} $\Act = \{ \act_0 \} \cup \{ \act_e \mid e \in E \}$ determine the movements.
%If agent $i$ moves next and its position is $(v,v',n)$ with $n > 0$, there is one unique action $\act_0$ at the corresponding state.
%That action has a unique successor state where $n$ is decremented by one and $i$ incremented modulo $m$, \ie, agent $i+1$ will move next.
%Thus, for $n>0$, there is no \textbf{decision} to be made.
%If $n = 0$, the agent has to decide which edge $(v',v'')$ to select.
%Intuitively, this selection means that the position of agent $i$ is set to $(v',v'',d(v',v''))$ and again $i$ is incremented modulo $m$.
% Edge selection has different types:
%If $i>0$ (an adversary moves next), there is a unique action $\act_0$ where the successor state is randomly determined according to the behavior $B_i$ for the current position of the adversary and the avatar.
%If $n = 0$ and $i=0$ (the avatar moves next), there is an action $\act_e$ reflecting every outgoing edge $e\in E$.
%% and the distributions are based on the uncertainty maps.
%Observe that the only states in which the avatar has to make choices are of the form $s_d=(v,v',0,\hdots,0)$.
%We call these states $s_d$ the \textbf{decision states}, from which the underlying decision procedure selects action $\act_e$ leading to a state $s_e =(v',v'', n, \hdots, 1)$.

\paragraph{Shield Construction.}
For
% an arena $(V,E,d)$ and the corresponding
the safety-relevant MDP $\mdp$, a set of unsafe states $T\subseteq S$ should preferably not be reached from any state.
The property $\varphi=\finally T$ encodes the \textbf{violation} of this safety constraint, that is, eventually reaching $T$ within $\mdp$.
The shield needs to limit the probability to \emph{satisfy} $\varphi$.
%The underlying property is $\varphi=\finally T$.
%If a certain probability to reach $T$ shall be ensured, we take a PCTL specification $\p_{\geq \lambda}(\varphi)$ with a lower bound $\lambda$ on the probability to satisfy $\varphi$.
%An example is $\p_{\geq 0.9}(\finally T)$, where the probability to reach the ``good'' target states $T$ is at least $90\%$.
We evaluate all decision states $s_d \in S_d$ with respect to this probability:
%We evaluate the states $s_e$ \wrt our safety specification $\varphi$ (a higher probability to satisfy $\varphi$ is better).
We compute $\mcresmin{\varphi}{\mdp}(s_e)$, i.e., the minimal probability to satisfy $\varphi$ from $s_e$, which is the state reached after taking action $\act_e\in\Act_e$ in $s_d$.
\begin{definition}[Action-valuation]\label{def:actionvalues}
	An \emph{action-valuation} for
%	each available
 	action	$\act_e\in\Act_e$
	at
%	each decision
	state $s_d\in S_d$ is
%	 given by
	\begin{align*}
 \val{s_d}{\mdp}\colon \Act(s_d) \rightarrow [0,1] \text{, with } \val{s_d}{\mdp}(\act_e) = \mcresmin{\varphi}{\mdp}(s_e)\ .
\end{align*}
The \emph{optimal action-value} for $s_d$ is $\optval{s_d}{\mdp} =  \min_{\act' \in \Act} \val{s_d}{\mdp}(\act')$, the set of all action-valuations at $s_d$ is $\actvalues_{s_d}$.
\end{definition}
We now define a shield for the safety-relevant MDP $\mdp$ using the action values.
Specifically, a \emph{$\delta$-shield} for $\delta\in[0,1]$ determines a set of  actions at each decision state $s_d$ that are $\delta$-optimal for the specification $\varphi$.
All other actions are ``shielded'' or ``blocked''.
% and cannot be chosen by a decision-maker.
%
\begin{definition}[Shield]\label{def:shield}
For action-valuation $\val{s_d}{\mdp}$ and $\delta \in [0,1]$, a \emph{$\delta$-shield for state $s_d\in S_d$} is
\begin{align*}
	\shield{\delta}{s_d}\colon & \actvalues_{s_d} \rightarrow 2^{\Act(s_d)}
\end{align*}
with $\shield{\delta}{s_d} \mapsto \{ \act \in \Act(s_d) \mid  \delta \cdot \val{s_d}{\mdp}(\act) \leq \optval{s_d}{\mdp} \}.$
%with $\shield{\delta}{s_d}(\val{s_d}{\mdp}) = \{ \act \in \Act(s_d) \mid \val{s_d}{\mdp}(\act) \geq \delta \cdot \optval{s_d}{\mdp} \}$.
\end{definition}
Intuitively, $\delta$ enforces a constraint on actions that are acceptable with respect to the optimal probability.
%\ie, which actions are $\delta$-optimal.
The shield is \emph{adaptive} with respect to $\delta$, as a high value for $\delta$ yields a stricter shield, a smaller value a more permissive shield.
The shield is stored using a lookup-table, and the value for $\delta$ can then be changed on-the-fly.
In particularly critical situations, the shield can enforce the decision-maker to resort to (only) the optimal actions w.r.t. the safety objective.

A $\delta$-shield for the MDP $\mdp$ is built by constructing and applying $\delta$-shields to all decision states.
\begin{definition}[Shielded MDP]\label{def:shield_mdp}
The \emph{shielded MDP} $\shielded{\mdp}=(S,\Act\,\shielded{\pmdp})$ for a safety-relevant quotient MDP $\MdpInitR$ and a $\delta$-shield for all $s_d\in S_d$ is given by the transition probability $\shielded{\pmdp}$ with
%The result $\shielded{\mdp} = \MdpTuple[']$ is a subMDP of $\mdp$ with
%\[S' = S, \quad \Act' = \bigcup \shield{\delta}{s}(\val{s}{\mdp}),  \text{ and }\quad
$\shielded{\pmdp}(s,\act) = \pmdp(s,\act)$ if $\act \in \shield{\delta}{s}(\val{s}{\mdp})$ and $\shielded{\pmdp}(s,\act) = \bot$ otherwise.
%\begin{align*}
%\shielded{\pmdp}(s,\act) = \begin{cases} \pmdp(s,\act) & \act \in \shield{\delta}{s}(\val{s}{\mdp})  \\
    %\bot & \text{otherwise.} \end{cases}	
%\end{align*}
\end{definition}
 %
%At least the actions that induce the optimal probability to satisfy $\varphi$ are always present in the shielded MDP.
%There is no state where all actions are blocked by the shield.
%
%We state the following formal properties of a shield.
\begin{lemma}
If MDP $\mdp$ is deadlock-free if and only if the shielded MDP $\shielded{\mdp}$ is deadlock-free.
%	For an MDP $\mdp$ and a $\delta$-shield, $\shielded{\mdp}$ is deadlock-free.
\end{lemma}
We compute the shield relative to optimal values $\optval{s_d}{\mdp}$.
Consequently, for $\delta=1$, only optimal actions are preserved, and for $\delta=0$ no actions are blocked.
%We assume that the original MDP $\mdp$ is deadlock-free.
%
\begin{theorem}\label{thm:shieldindependent}
For an MDP $\mdp$ and a $\delta$-shield, it holds for any state $s$ that
$\val{s}{\mdp} = \val{s}{\shielded{\mdp}}$.
\end{theorem}
As optimal actions for the safety objective are not removed, optimality w.r.t.\ safety is preserved in the shielded MDP.
%Moreover, if an action is allowed by the shield, all probabilistic outcomes as described by $\pmdp$ are preserved.
Thus, during construction of the shield, we compute the action-valuations in fact \emph{for the shielded MDP}.
Observe that computing a shield for a state is done \emph{independently} from the application of the shield to other states.

%In particular, computing a shield for a state is \textbf{independent from the application of the shield to other states}.
%Consequently, allowing $\delta-$optimal actions at $s$ does not affect the $\delta-$optimality of actions in $s'$.

\paragraph{Guaranteed Safety.}
A $\delta$-shield ensures that only actions that are $\delta$-optimal with respect to an LTL property $\varphi$ are allowed.
In particular, for each action $\act\in\Act_e$ at state $s_e$, we use the \emph{minimal} probability $\mcresmin{\varphi}{\mdp}(s_e)$ to satisfy $\varphi$, see Def.~\ref{def:actionvalues}.
Under \emph{optimal} (subsequent) choices, the value $\mcresmin{\varphi}{\mdp}(s_e)$ will be achieved.
In contrast, a sequence of bad choices may violate $\varphi$ with high probability.
A more conservative notion would be to use the minimal action value while assuming that in all subsequent states the worst-case decisions corresponding to the maximal probabilities are taken.
% $\val{s_d}{\mdp}(\act_e)=\mcresmax{\varphi}{\mdp}(s_e)$
%as action-valuations.
These values are computable by model checking.
Regardless of subsequent choices, at least $\val{s_d}{\mdp}(\act_e)$ is then guaranteed.
A sensible notion to construct a shield would then be to impose a threshold $\lambda\in[0,1]$ such that only actions with $\val{s_d}{\mdp}(\act_e)\leq\lambda$ are allowed.
A shield with such a guaranteed safety probability may induce a shielded MDP (Def.~\ref{def:shield_mdp}) that is \emph{not deadlock free}.
Moreover, the shield may become too restrictive for the agent.

\paragraph{Scalable Shield Construction.}\label{sec:faster}
Although we apply model checking only in the  safety-relevant MDP, scalability issues for large applications remain.
We employ several optimizations towards computational tractability.

\emph{Finite Horizon.}
% (the arena).
For infinite horizon properties, the probability to violate safety (in the long run) is often one.
Furthermore, our learned MDP model is inherently an approximation of the real world.
 Errors originating from this approximation accumulate for growing horizons. %and the probability of reaching critical states may be high.
Thus, we focus on a finite horizon such that the action values (and consequently, a policy for the avatar) carry only guarantees for the next steps.
This assumption also allows us to prune the
%relevant fragment of the
safety-relevant MDP (see below), increasing the scalability.

%Technically, that is true if the underlying graph of the MDP is fully connected.

\emph{Piecewise Construction.}
Computing a shield for all states in an MDP concurrently yields a large memory footprint.
To alleviate this footprint, we compute the shield states independently, in accordance with Theorem~\ref{thm:shieldindependent}.
The independent computation prunes the relevant part of the MDP, as the number of states reachable within the horizon is drastically reduced.
Additionally, the independent computation allows for parallelizing the computation.

\emph{Independent Agents.}
The explosion of state spaces stems mostly from the number of agents.
Here, an important observation is that we can consider agents independently.
For instance, the probability for the avatar to crash with an adversary is stochastically independent from crashing with the others.
Instead of determining the shield for all adversaries at once, we perform computations for each agent individually, and combine them via the inclusion-exclusion principle.
Afterwards, the shield is composed from the shields dedicated to individual adversaries.

\emph{Abstractions.}
We observe that for finite horizon properties and piecewise construction, adversaries may be far away---beyond the horizon---without a chance to reach the avatar.
We do not need to consider such (positions for) adversaries,
%To remove redundancy induced by considering states independently, we pre-analyze that in some states all actions allow for a high probability to satisfy the safety specification.
as in these states, the shield will not block any actions.
%Consequently, we omit model checking.
%In combination with finite horizon properties the shield still guarantees safety in each state exactly for the specified horizon.
%
%In our computation, we neglect all adversary positions that are not relevant at the current state.
%Such states, which indeed violate the safety specification with probability $0$, are excluded from the state space prior to model checking.
%For finite-horizon properties, they may not even be part of the MDP we build.

%
%Not good to iterate over all policies, too many.
%However, many policies only differ 'beyond the horizon' when considered from a particular starting point.
%This horizon is known during model construction.
%Number of different policies is then easily manageable.
%
%
%Cut off (before model checking)! states with prob 0. If agents are too far apart, they cannot crash within horizon distance. Thus, do not affect measure that the model checker considers.
%
%Agents are anonymous, in the model they behave the same except for order of movement.
%Order of movement has no effect, as agents move independently.
%Thus, apply symmetry reduction (by reordering the index of the agent based on there relative position).

{\emph{Fewer Decision States.}
Depending on the setting, there might be only a few critical situations in which the agent requires shielding to ensure safety. The shield can be computed for this critical
states only. Consequently, the agent makes shielded decisions in the adapted decision states, and unshielded decisions in all other ones.

\paragraph{Shielding versus Performance.}\label{sec:progress}
A shield which is \emph{minimally invasive} gives the RL agent the most freedom to optimize the performance objective.
We propose two methods to alleviate invasiveness, all of them assume \emph{domain knowledge} of the rationale behind the decision procedure.
%\textcolor{green}{A third option is detailed in~\cite[Section 1.2]{supp}}.
% \eg in the form of a measure $\progress\colon S \rightarrow [0,1]$.

\emph{Iterative Weakening.}
During runtime, we may observe that the progress of the avatar regarding the performance objective is not increasing anymore.
Then, we weaken the shield by $\delta-\varepsilon$, allowing additional actions.
As soon as progress is made, we reset $\delta$ to its former value.
%We still guarantee $\delta-\varepsilon$ safety overall.
The adaption of $\shield{\delta}{s}$ to $\shield{\delta - \varepsilon}{s}$ can be done on the fly, without new computations.
%On-the-fly has downside that it is not clear upfront when the guarentees will be lowered.

\emph{Adapted Specifications.}
If the goal of the decision maker is known \emph{and} can be captured in temporal logic, we may adapt the original specification accordingly.
There are often natural trade-offs between safety and performance.
These trade-offs might be resolved via weights, but this process is often undesirable~\cite{DBLP:journals/jair/RoijersVWD13} and similar to reward engineering.
Instead, optimizing the conditional performance while assuming to stay sufficiently safe \cite{DBLP:conf/aaai/Teichteil-Konigsbuch12}, avoids side-effects of attaching some weights to the safety specification.
%\begin{example}
%\sj{Challenge: make the correct types of combined specs}
%
%Conditional $ \finally\{ s' \mid \progress(s') = 1 \} \mid \varphi$
%For simple safety criterions (such as ``do not crash``) the specification can be simplified to
%$\varphi \pctlUntil \{ s' \mid \progress(s') = 1 \}$.
%As the horizon for to reach $\progress(s')$ might be very large, a typical approximation would be to just require safety until \emph{some progress}, i.e. when starting from state $s$
%$\varphi \pctlUntil \{ s' \mid \progress(s') > \progress(s) \}$.
%
%\end{example}
%Safety is not optimal here, safety might still be higher if taking more progress at slightly higher risk.
%
%Often, it is necessary to also extend the arena:
%The goal of the avatar might include actions and states that are irrelevant for the safety specification.
%
%Combination of larger arena and more complex properties is challenging.

%
%
\section{Implementation and Numerical Experiments}\label{sec:experiments}
\pgfplotsset{every tick label/.append style={font=\tiny}}
%We discussed a framework (1) to learn stochastic behavior models for adversaries, (2)  to construct a shield for an MDP, (3) to enable the computationally tractable construction of such a shield, and (4) to provide sufficient progress for a shielded learning algorithm.
%
%We translate an arena and a position of the avatar into a relevant fragment of the MDP. We describe the MDP in the high-level \prism-language.
%

%To demonstrate our methods, we consider the arcade game \pacman.
%\subsection*{Applications}

\paragraph{Set-up.}
We run experiments using an Intel Core i7-4790K CPU with 16 GB of RAM using 4 cores.
We give the timing results for a single CPU. Since the shield may be computed in a multi-threaded architecture, this time can be divided by the number of cores available.

The supplementary materials, namely the source code and videos are available online\footnote{\url{http://shieldrl.nilsjansen.org}}.

We demonstrate the applicability of our approach by means of two case studies.
For both case studies, we learn adversary behavior in small arenas individually for each adversary.
These behavior models are applicable to any benchmark instance, as they are independent of concrete positions.

\begin{figure}[t]
	\centering
%	\subfigure[Small \pacman]
%	{
%	 \raisebox{0.75\height}{\includegraphics[scale=0.07,bb= 10 0 1900 510]{pics/vido_pacman_traps.png}}
%	\label{fig:video_small}
%	}
%	\qquad
%	\subfigure[Resulting Scores for small \pacman]
%	{
%      \scalebox{0.85}{
%	  \begin{tikzpicture}
%      \begin{axis}[
%        legend style={at={(0.97,0.45)},anchor=east},
%        width=7cm,height=5cm,
%        grid=major,
%        ymax=1000,
%        xlabel=Training Episodes,
%        ylabel=Average Reward,
%        xtick={0,40,80,120,160,200,240,280}
%        ]
%        \addplot[mark=*, blue, solid] table[x=episodes,y=woshield] {datasets/small.dat};
%        \addlegendentry{Without Shield}
%
%        \addplot[mark=square*, orange, densely dashed]  table[x=episodes,y=wshield]  {datasets/small.dat};
%        \addlegendentry{With Shield}
%      \end{axis}
%      \end{tikzpicture}
%      \label{fig:result_small}
%	  }
%    }
	\subfigure[Still from video on classic PAC-MAN]
	{
	 %\raisebox{0.1\height}{
\includegraphics[scale=0.23,bb= 10 0 1000 510]{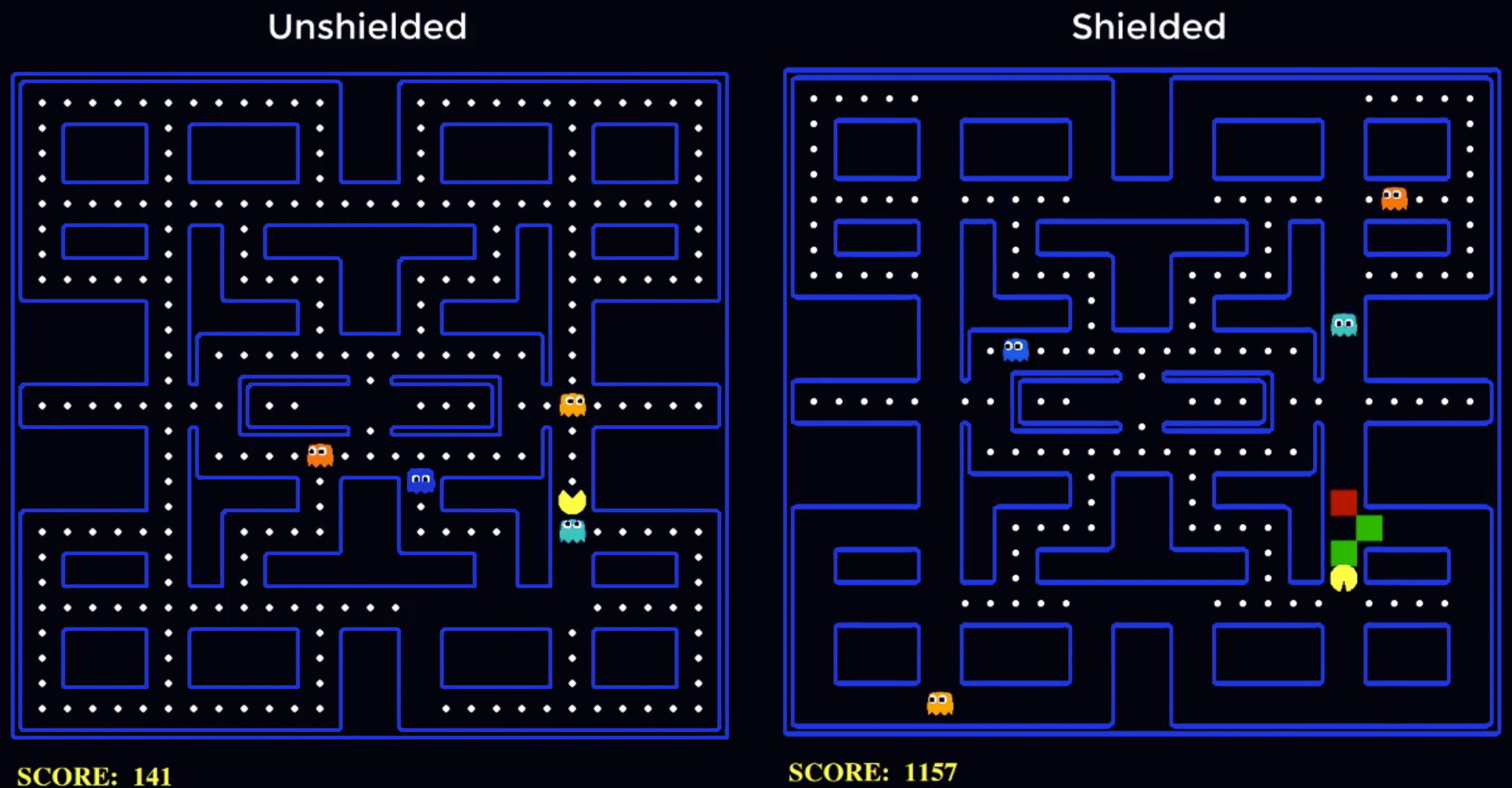}
%}
	\label{fig:video_classic}
	}
	%\qquad
	\subfigure[Scores during training for classic PAC-MAN]
	{
	  \scalebox{1}{
	  \begin{tikzpicture}
      \begin{axis}[
        legend style={font=\scriptsize, at={(1.24,1)},anchor=north}, legend columns=1,
        width=6cm,height=5cm,
        grid=major,
        ymax=1000,
        xlabel=Training episodes,
        ylabel=Average reward,
        xtick={0,40,80,120,160,200,240,280},
        mark options={scale=0.6}
        ]
        [

        width=6cm,height=5cm,
        grid=major,
        ymax=750,
        ytick={-500,0,500,750},
        xlabel=Training episodes,
        ylabel=Average reward,
        xtick={0,40,80,120,160,200,240,280},
        mark options={scale=0.6}
        ]

        \addplot[mark=*, blue, solid] table[x=episodes,y=woshield] {datasets/classic.dat};
        \addlegendentry{W/o Shield}

        \addplot[mark=square*, orange, densely dashed]  table[x=episodes,y=wshield]  {datasets/classic.dat};
        \addlegendentry{With Shield}
      \end{axis}
      \end{tikzpicture}

	  }
\label{fig:result_classic}
	}%new removed

\label{fig:pacmantable}
	\caption{Scenarios and results for PAC-MAN}
\end{figure}
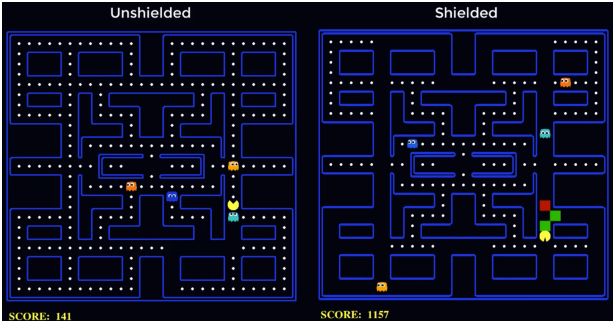
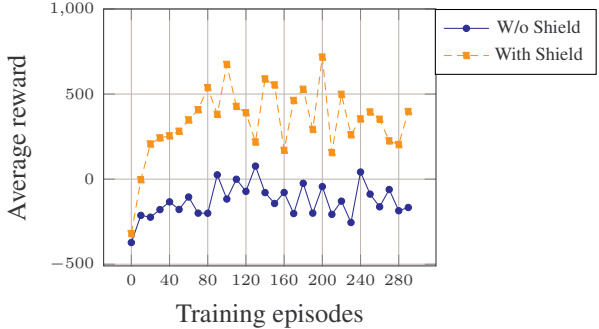

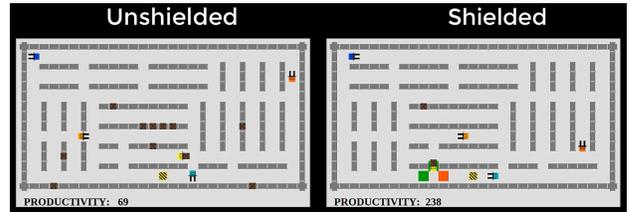
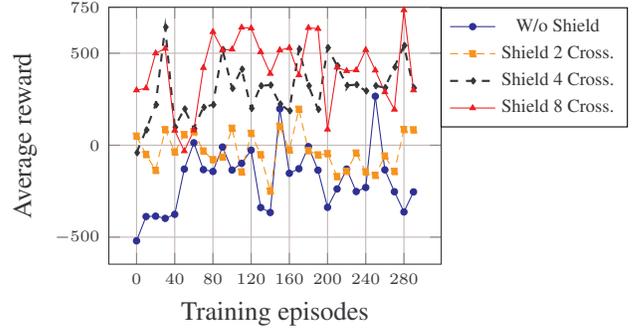
\begin{figure}[t]
%	\centering
	%\raisebox{2cm}{
	\subfigure[Still from the video on warehouse]{
	
\includegraphics[scale=0.095]{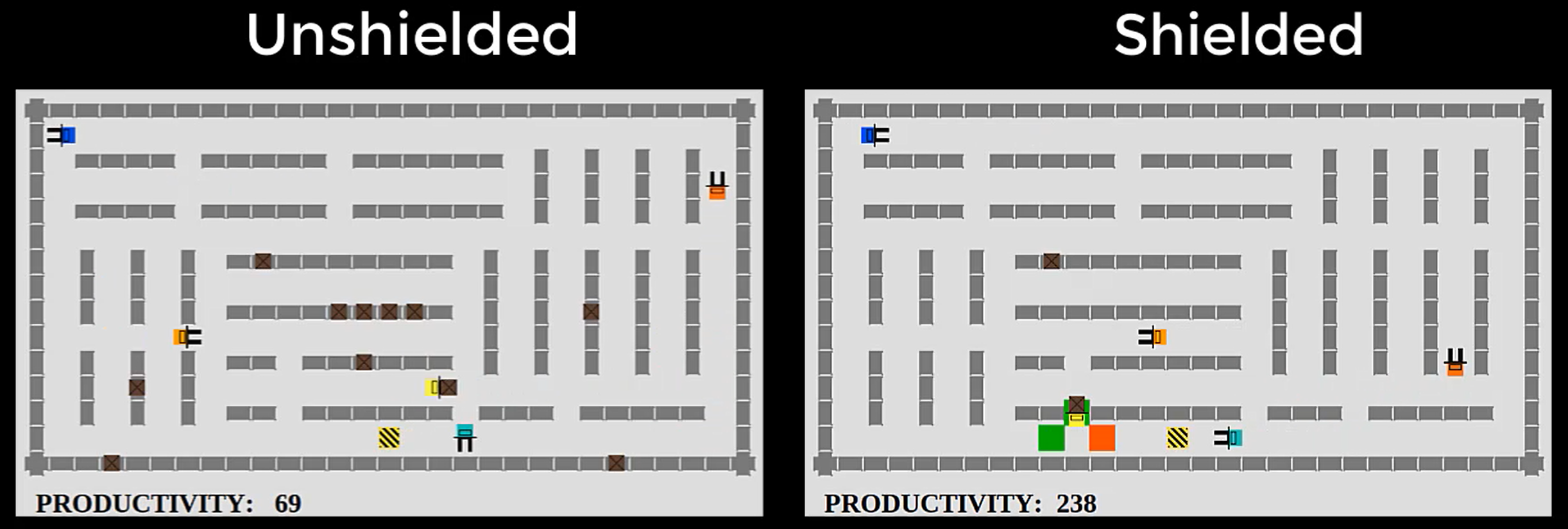}
\label{fig:warehousevid}
}
%}
\subfigure[Scores during training on warehouse]
	{
	  \scalebox{1}{%{0.85}
	  \begin{tikzpicture}
      \begin{axis}[
      legend style={font=\scriptsize, at={(1.28,1)},anchor=north}, legend columns=1,
        width=6cm,height=5cm,
        grid=major,
        ymax=750,
        ytick={-500,0,500,750},
        xlabel=Training episodes,
        ylabel=Average reward,
        xtick={0,40,80,120,160,200,240,280},
        mark options={scale=0.6}
        ]
        \addplot[mark=*, blue, solid] table[x=episodes,y=woshield] {datasets/warehouse.dat};
        \addlegendentry{W/o Shield}

        \addplot[mark=square*, orange, densely dashed]  table[x=episodes,y=shield2]  {datasets/warehouse.dat};
        \addlegendentry{Shield 2 Cross.}

        \addplot[mark=diamond*, black, dashed, thick]  table[x=episodes,y=shield4]  {datasets/warehouse.dat};
        \addlegendentry{Shield 4 Cross.}

        \addplot[mark=triangle*, red, solid]  table[x=episodes,y=shield8]  {datasets/warehouse.dat};
        \addlegendentry{Shield 8 Cross.}

      \end{axis}
      \end{tikzpicture}
      	  }
\label{fig:warehousegraph}
}

\caption{Scenarios and results for warehouse}
\end{figure}

%\paragraph{PAC-MAN.}
For the arcade game PAC-MAN, \pacman (the avatar) aims to collect \emph{PAC-dots} in a \emph{maze} and not get caught by \emph{ghosts} (the adversaries).
We model various instance of the game (with different sizes) as an arena, where tokens represent the dots at each position in the maze, such that a dot is either present or collected.
The score (reward, performance) is positively affected (+10) by collecting a dot and negatively by time (each step: -1). If \pacman either collects all dots (+500) or is caught (-500), the game is restarted.
RL approaches exist~\cite{pacman}, but they suffer from the fact that during the exploration phase \pacman is often caught by the ghosts, achieving very poor scores.
The safety specification places a lower bound on the probability of reaching states in the underlying MDP that correspond to being caught.
%Food earns reward ($+10$), while each step causes a small penalty ($-1$).
%A large reward ($+500$) is granted, if \pacman eats all the food in the maze.
%If \pacman gets eaten, a large penalty ($-500$) is imposed and the game is restarted.

%\paragraph{Fork-lift robot in a warehouse.}
We also consider a warehouse floor plan with several corridors.
A similar scenario has been investigated in~\cite{DBLP:journals/corr/abs-1806-07135}.
In the arena, nodes describe crossings, the edges the corridors with shelves, and the distances the corridor length.
The agents are fork-lift units picking up packages from the shelves and delivering them to the exit; tokens represent the presence of a package at its position.
The avatar is a specific (yellow) fork-lift unit that has to account for other units, the adversaries.
The performance (reward) is positively affected by loading and delivering packages (+20, respectively) and negatively by time (each step: -1).
Delivering all packages yields a large bonus (+500) and a collision leads to a large punishment (-500), both cases end the scenario.
%
%
%inspecting machines where an issue has been raised (as indicated by a token), while accounting for the behavior of the adversaries.
% All agents follow the corridors and take another corridor upon reaching a crossing.
 Corridors might be too narrow for multiple (facing) units, which poses a safety-critical situation.
Most crucial is the crowded area near the exit, since all units have to deliver the packages to the exit.
Transferring the stochastic adversary behavior to any arena (without tokens) yields a concrete safety-relevant MDP.
In particular, we specify an arena with the positions of the avatar and the adversaries as well as the behavior in the high-level \prism-language~\cite{DBLP:conf/cav/KwiatkowskaNP11}.
We employ a script that automatically generates arenas to enable a broad set of benchmarks.
Taking, e.g., the PAC-MAN arena from Fig.~\ref{fig:video_classic}, the considered MDP has roughly $10^{12}$ states (compared to $10^{50}$ for the full MDP).
For a safety-relevant MDP, we compute a $\delta$-shield (with iterative weakening) via the model checker \storm~\cite{DJKV17}, using a horizon of $10$ steps.
The immense size even of safety-relevant MDPs requires optimizations such as a piecewise and independent shield construction.
Moreover, a multi-threaded architecture lets us construct shields for very large examples.
In particular, we perform model checking for (many) MDPs of roughly $10^6$ states.
The computation time for the largest PM instance takes about 6 hours (single-threaded), while memory is not an issue due to the piecewise shield construction.

%\paragraph{Reinforcement Learning.}
We compare RL to shielded RL on different instances.
The key comparison criterion is the performance (detailed above) during learning.
Our implementation is based on an existing PAC-MAN environment\footnote{\url{http://ai.berkeley.edu/project_overview.html}} using an approximate Q-learning agent~\cite{sutton1998reinforcement}
with the following feature vectors:
\begin{compactitem}
\item for PAC-MAN: (1)~distance to the closest dot, (2)~whether a ghost collision is imminent, and (3)~whether a ghost is one step away.
\item for Warehouse: (1)~has the unit loaded or unloaded, (2)~the distance to the next package and (3)~to the exit, (4)~whether another unit is three steps away and (5)~one step away.
\end{compactitem}
The results are basic reflex controllers.
The Q-learning uses the learning rate $\alpha = 0.2$ and the discount factor $\gamma=0.8$
for the Q-update and an $\epsilon$-greedy exploration policy with $\epsilon=0.05$.
One episode lasts until either the game is restarted.
We describe results for the training phase of RL (300 episodes).
%
%\textbf{BK note: Here are the Q-Learning Details for Warehouse:}
%\begin{itemize}
%  \item Features:
%  \item 1) has the unit loaded or unloaded
%  \item 2) how far away the next package is
%  \item 3) how far away the exit is
%  \item 4) whether another unit is one step away
%  \item 5) whether another unit is three step away
% % \item 6) Reward: +20 load a package, +20 unload the package at exit, -1 per step, +500 deliver all packages
%
%\end{itemize}

%
%
%number of steps needed to eat all food and by the number of times \pacman gets eaten.\nj{score may be redundant here.}

%\begin{figure}[t]
%	\centering
%	\includegraphics[scale=0.47]{pics/tabledummy.png}
%	\label{tab:table}
%	\caption{results}
%\end{figure}
% \usepackage{booktabs}

\paragraph{Results.}
Figures~\ref{fig:video_classic} and~\ref{fig:warehousevid} show screenshots of a series of recommended videos (available in the supplementary material).
Each video compares how RL performs either shielded or unshielded on a instance of the case study.
In the shielded version, at each decision state in the underlying MDP, we indicate the risk of decisions from low to high by the colors green, orange, red.
%This feature also enables safe game-playing for humans rather than RL based agents.

%
Consider PAC-MAN in detail:
Figure~\ref{fig:result_classic} depicts the scores obtained during RL.
The curves (blue, solid: unshielded, orange, dashed: shielded) show the average scores for every ten training episodes.
Table~\ref{tab:table1} shows results for instances in increasing size.
We list the number of model checking calls and the time to construct the shield.
We list the scores with and without shield, and the \emph{winning rate} capturing the ratio of successfully ended episodes.
For all instances, we see a large difference in scores due to the fact that \pacman is often rescued by the shield.
The winning rates differ for most benchmarks, favoring shielded RL.
For three or four ghosts, a shield with a ten-step horizon cannot guide \pacman to avoid being encircled by the ghosts long enough to successfully end the game.
Nevertheless, the shield often safes \pacman, leading to superior scores.
Moreover, the shield helps learning an optimal policy much faster as fewer restarts are needed.
%More results are depicted in~\cite[Section 1.3]{supp}.

For the warehouse case study, we choose to vary the decision states, i.e., the positions of the avatar for which we compute a shield. We present results for shielding the 2--8 crossings closest to the exit.
Figure~\ref{fig:warehousegraph} shows the average score  for the different variants, Table~\ref{fig:warehousetable} summarizes average score and win rate.
Unsurprisingly, the score gets better the more states are shielded. Furthermore, we have seen that shielding even more states has only a very limited effect.

\begin{table}[t]
\scriptsize
\def\arraystretch{0.7}%  1 is the default, change whatever you need
\setlength{\tabcolsep}{0.45em} % for the horizontal padding
\centering
\caption{Average scores and win rates for \pacman}
\label{tab:table1}
\begin{tabular}{@{}|r|r r|r r|r r|@{}}
\toprule
\multicolumn{1}{|c|}{\textbf{\begin{tabular}[c]{@{}c@{}}Size,\\ \#Ghosts\end{tabular}}} & \multicolumn{1}{c}{\textbf{\begin{tabular}[c]{@{}c@{}}\#Model \\ Checking\end{tabular}}} & \multicolumn{1}{c|}{\textbf{time (s)}} & \multicolumn{1}{c}{\textbf{\begin{tabular}[c]{@{}c@{}}Score\\  w/o Shield\end{tabular}}} & \multicolumn{1}{c|}{\textbf{\begin{tabular}[c]{@{}c@{}}Score w. \\ Shield\end{tabular}}} & \multicolumn{1}{c}{\textbf{\begin{tabular}[c]{@{}c@{}}Win Rate \\ w/o Shield\end{tabular}}} & \multicolumn{1}{c|}{\textbf{\begin{tabular}[c]{@{}c@{}}Win Rate \\ w. Shield\end{tabular}}} \\ \midrule
9x7,1                                                                                   & 5912                                                                                                & 584                                   & -359,6                                                                                    & 535,3                                                                                     & 0,04                                                                                         & 0,84
\\ \midrule
17x6,2                                                                                  & 5841                                                                                                & 1072                                     & -195,6                                                                                    & 253,9                                                                                      & 0,04                                                                                         & 0,4
\\ \midrule
17x10,3                                                                                 & 51732                                                                                               & 3681                                   & -220,79                                                                                   & -40,52                                                                                     & 0,01                                                                                         & 0,07
\\ \midrule
27x25,4                                                                                 & 269426                                                                                              & 19941                                   & -129,25                                                                                   & 339,89                                                                                     & 0,00                                                                                          & 0,00
\\ \bottomrule
\end{tabular}
\end{table}
\normalsize

\begin{table}[t]
\scriptsize
\centering
\caption{Average scores and win rates for warehouse}
\label{fig:warehousetable}
\begin{tabular}{@{}|l|rrrr|@{}|}
\toprule
\textbf{Crossings shielded}         & \textbf{0} & \textbf{2} & \textbf{4} & \textbf{8} \\\midrule
Score    & -186       & -27.6               & 303               & 420                  \\
Win Rate & 0.16       & 0.31                 & 0.59                 & 0.71
\\\bottomrule
\end{tabular}
\end{table}

%\paragraph{Discussion.}
%\sj{rephrase}
%In general, learning for an arcade game like \pacman is difficult to perform according to safety constraints if no knowledge about future events is available.
%Given our relatively loose assumptions about the setting, the shield proved a feasible means to ensure an appropriate measure of safety.
%Moreover, as in the classic \pacman instance, 100\% safety is not possible.\nj{what about warehouse here?}
%\vspace{-0.3cm}
%In our experiments, we found that while finite horizons of $10$ steps lead to good results, in several cases (such as long hallways), presumably safe behavior was in fact unsafe.

%
%\section{Discussion and related work}
%
\section{Conclusion and Future Work}\label{sec:conclusion}
We developed the concept of shields for MDPs. 
Utilizing probabilistic model checking, we maintained probabilistic safety measures during reinforcement learning.
We addressed inherent scalability issues and provided means to deal with typical trade-off between safety and performance.
Our experiments showed that we improved the state-of-the-art in safe reinforcement learning.

For future work, we will extend shields to richer models such as partially-observable MDPs. 
Moreover, we will extend the applications to more arcade games and employ deep recurrent neural networks as means of decision-making~\cite{hausknecht2015deep,DBLP:conf/ijcai/Carr0WS0T19}.
Another interesting direction is to explore (possibly model-free) learning of shields, instead of employing model-based model checking.

%\newpage
%\pagebreak

\begin{small}
\bibliographystyle{aaai}
\bibliography{literature}
\end{small}

%\newpage
%\input{appendix}

%%
\end{document}